\documentclass[11pt,a4paper]{article}
\usepackage[hyperref]{emnlp2018}
\usepackage{times}
\usepackage{latexsym}
\usepackage{url}

\aclfinalcopy 

\usepackage{helvet}
\usepackage{courier}
\usepackage{graphicx}
\usepackage{enumitem}

\usepackage{latexsym}
\usepackage{amsmath,amsfonts}
\usepackage{mathtools}
\usepackage{algorithm2e}
\usepackage{color}
\usepackage{url}
\usepackage{mathtools}
\usepackage{pgf}
\usepackage{multirow}
\usepackage{soul}

\usepackage{microtype}

\newenvironment{itemizesquish}[2]{\begin{list}{\labelitemi}{\setlength{\itemsep}{#1}\setlength{\labelwidth}{#2}\setlength{\leftmargin}{\labelwidth}\addtolength{\leftmargin}{\labelsep}}}{\end{list}}

\newcommand{\sebastiao}[1]{\textcolor{black}{#1}} 

\title{Multilingual Clustering of Streaming News}

\author{Sebasti\~{a}o Miranda$^{\diamondsuit}$ \quad Art\={u}rs Znoti\c{n}{\v s}$^{\dagger\star}$ \quad Shay B. Cohen$^{\heartsuit}$ \quad Guntis Barzdins$^{\dagger\star}$ \smallskip \\
$^{\diamondsuit}$Priberam Labs, Alameda D.~Afonso Henriques, 41, 2$^{\mathrm{o}}$, 1000-123 Lisboa, Portugal \\
$^{\dagger}$Innovation Labs LETA, Marijas Str. 2, Riga LV-1050, Latvia \\
$^{\star}$University of Latvia, IMCS, Rainis Blvd. 29, Riga LV-1459, Latvia  \\
$^\heartsuit$School of Informatics, University of Edinburgh, Edinburgh EH8 9AB \smallskip \\
{\tt sebastiao.miranda@priberam.pt}, {\tt arturs.znotins@leta.lv}, \\
{\tt scohen@inf.ed.ac.uk}, {\tt  guntis.barzdins@lu.lv}
}

\newcommand{\bottom}{\bot}

\newcommand{\ignore}[1]{}

\begin{document}
	
	\maketitle
	
	\begin{abstract}
		Clustering news across languages enables efficient media monitoring by aggregating articles from
		multilingual sources into coherent stories. Doing so in an online setting allows
		scalable processing of massive news streams. To this end, we describe a novel method for clustering an incoming
		stream of multilingual documents into monolingual and crosslingual story clusters. Unlike typical clustering approaches that consider a small and known number of labels, we tackle the problem of discovering an ever growing number of cluster labels in an online fashion, using real news datasets in multiple languages. 
		Our method is simple to implement, computationally efficient and produces state-of-the-art results on datasets in German, English and Spanish.\\  
	\end{abstract}
	
	\section{Introduction}
	Following developing news stories is imperative to making real-time decisions on important political and public safety matters. Given the abundance of media providers and languages, this endeavor is an extremely difficult task. As such, there is a strong demand for automatic clustering of news streams, so that they can be organized into stories or themes for further processing. Performing this task in an online and efficient manner is a challenging problem, not only for newswire, but also for scientific articles, online reviews, forum posts, blogs, and microblogs.
	
	A key challenge in handling document streams is that the story clusters must be generated on the fly in an online fashion: this requires handling documents one-by-one as they appear in the document stream. In this paper, we provide a treatment to the problem of {\em online document clustering}, i.e. the task of clustering a stream of documents into themes. For example, for news articles, we would want to cluster them into related news stories. 	
	
	To this end, we introduce a system which aggregates
	news articles into fine-grained story clusters across
	different languages in a completely online and scalable
	fashion from a continuous stream. Our clustering approach is part of a larger {\em media monitoring} project to solve the problem of monitoring massive text and TV/Radio streams (speech-to-text). In particular, media monitors write intelligence reports about the most relevant events, and being able to search, visualize and explore news clusters assists in gathering more insight about a particular story. Since relevant events may be spawned from any part of the world (and from many multilingual sources), it becomes imperative to cluster news across different languages. 
	
	In terms of granularity, the type of story clusters we are interested in are the group of articles which, for example : (i) Narrate recent air-strikes in Eastern Ghouta (Syria); (ii) Describe the recent launch of Space X's Falcon Heavy rocket.

	\paragraph{Main Contributions}
	
	While most existing news clustering approaches assume a monolingual document stream -- a non-realistic scenario given the diversity of languages on the Web -- we assume a general, multilingual, document stream. This means that in our problem-formulation story documents appear in multiple languages and we need to cluster them to crosslingual clusters. 
	Our main contributions are as follows:

	\begin{itemizesquish}{-0.3em}{0.5em}
	
		\item We develop a system that aggregates
	news articles into fine-grained story clusters across
	different languages in a completely online and scalable
	fashion from a continuous stream. As discussed in the introduction, this is a highly relevant task for the use-case of media monitoring.
		
	\item We formulate the problem of online multilingual document clustering and the representation that such clustering takes by interlacing the problem of monolingual clustering with crosslingual clustering. The representation of our clusters is interpretable, and similarly to topic models, consists of a set of keywords and weights associated with the relevant cluster. In our formulation, a monolingual cluster is a group of documents, and a crosslingual cluster is a group of monolingual clusters in different languages.
	
	\item We compare our approach to our own implementation of a state-of-the-art streaming method, and show much superior results for a dataset in English, Spanish and German.
	
	\end{itemizesquish}
	
	\section{Problem Formulation}
\vspace{-0.1in}

	We focus on clustering of a stream of documents, where the number of clusters is not fixed and learned automatically. We denote by $D$ a (potentially infinite) space of multilingual documents. Each document $d$ is associated with a language in which it is written through a function $L \colon D \rightarrow \mathcal{L}$ where $\mathcal{L}$ is a set of languages. For example, $L(d)$ could return English, Spanish or German. (In the rest of the paper, for an integer $n$, we denote by $[n]$ the set $\{ 1, \ldots, n\}$.)
	
	We are interested in associating each document with a monolingual cluster via the function $C(d) \in \mathbb{N}$, which returns the cluster label given a document. This is done independently for each language, such that the space of indices we use for each language is separate.
	
	Furthermore, we interlace the problem of monolingual clustering with crosslingual clustering. This means that as part of our problem	formulation we are also interested in a function $E \colon \mathbb{N} \times \mathcal{L} \rightarrow \mathbb{N}$ that associates each monolingual cluster with a crosslingual cluster, such that each crosslingual cluster only groups one monolingual cluster per different language, at a given time. The crosslingual cluster for a document $d$ is $E(C(d), L(d))$. As such, a crosslingual cluster groups together monolingual clusters, at most one for each different language.
	
	Intuitively, building both monolingual and crosslingual clusters allows the system to leverage high-precision monolingual features (e.g., words, named entities) to cluster documents of the same language, while simplifying the task of crosslingual clustering to the computation of similarity scores across monolingual clusters - which is a smaller problem space, since there are (by definition) less clusters than articles. We validate this choice in \S\ref{section:experiments}.
		
	\section{The Clustering Algorithm}
	\label{section:alg}
	
    Each document $d$ is represented by two vectors in $\mathbb{R}^{k_1}$ and $\mathbb{R}^{k_2}$.
	The first vector exists in a ``monolingual space'' (of dimensionality $k_1$) and is based on a bag-of-words representation of the document. The second vector exists in a ``crosslingual space'' (of dimensionality $k_2$) which is common to all languages. More details about these representations are discussed in \S\ref{sec:doc}.

	\paragraph{Online Clustering}
	With our clustering algorithm, we maintain two types of centroid functions for each monolingual cluster.
	The first is a centroid function $H \colon \mathbb{N} \times \mathcal{L} \rightarrow \mathbb{R}^{k_1} \cup \{ \bottom \}$ that assists in associating
	each document with a monolingual cluster. The second is a centroid function $G \colon \mathbb{N} \rightarrow \mathbb{R}^{k_2} \cup \{ \bottom \}$ that
	assists in associating each monolingual cluster with a crosslingual cluster. The $\bottom$ symbol is reserved to denote documents which are not associated with any cluster yet.
	
	In our algorithm, we need to incrementally construct the functions $H$, $G$ (the two centroid functions), $C$ (the monolingual clustering function)
	and $E$ (the crosslingual clustering function). Informally, we do so by first identifying a monolingual cluster for an incoming document by finding
	the closest centroid with the function $H$, and then associate that monolingual cluster with the crosslingual cluster that is closest based on the function
	$G$. The first update changes $C$ and the second update changes $E$. Once we do that, we also update $H$ and $G$ to reflect the new information that	exists in the new incoming document.

\begin{figure}[t]
		\centering
		\includegraphics[width=3in]{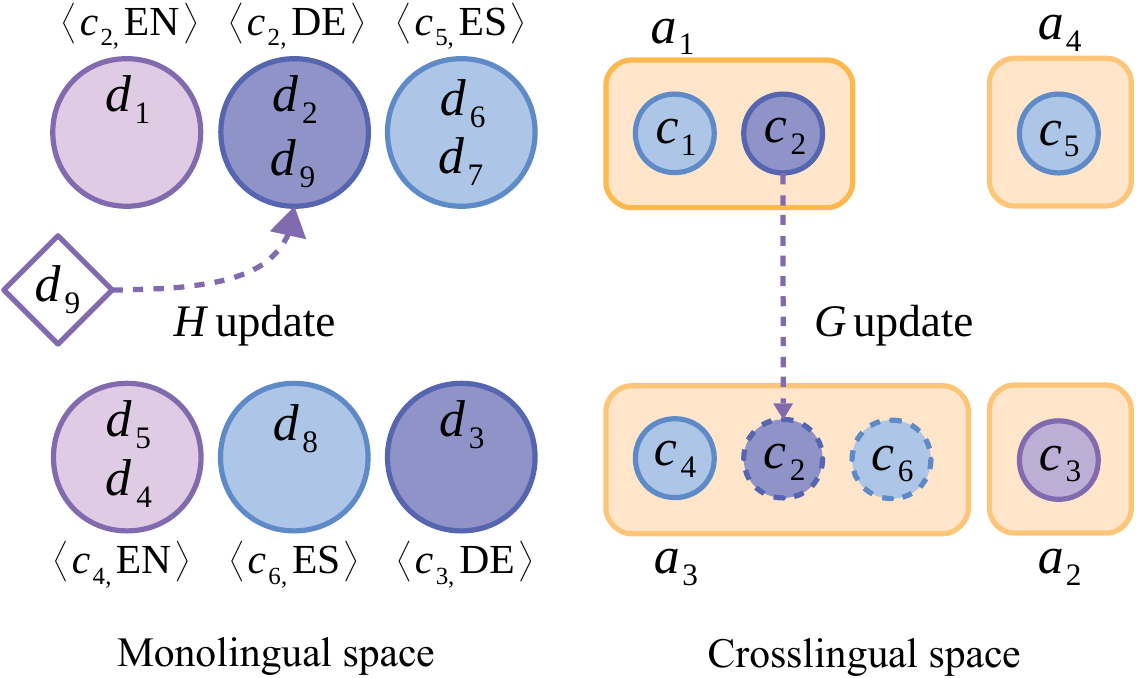}
		
		\caption{\label{fig:alg} A pictorial description of the algorithm and the state it maintains. The algorithm maintains a monolingual cluster space,
in which each cluster is a set of documents in a specific language. The algorithm also maintains a crosslingual cluster space, in which a cluster is a set
of monolingual clusters in different languages. Documents are denoted by $d_i$, monolingual clusters by $c_i$ (circles) and crosslingual clusters by $a_i$.}

	\end{figure}

	\paragraph{Example} Figure~\ref{fig:alg} depicts the algorithm and the state it maintains. A document in some language ($d_9$) appears in the stream, and is clustered into one of the monolingual clusters (circles) that group together documents about the same story (for example, $\langle c_2, \mathrm{DE} \rangle$ could be a German cluster about a recent political event). Then, following this monolingual update, the online clustering algorithm updates the crosslingual clusters (round rectangles), each grouping together a set of monolingual clusters, one per language at the most. The centroids for the monolingual clusters are maintained by the function $H$. For example, $H(2, \mathrm{English})$ gives the centroid of the upper left English monolingual cluster. The function $G$ maintains the crosslingual clusters. Considering the upper-left most crosslingual cluster, $a_1$, then $G(1)$ returns its centroid.

	\paragraph{Algorithm} To be more precise, the online clustering process works as follows. $H$ and $G$ start with just returning $\bottom$ for any cluster number, both monolingual and crosslingual.
	With a new incoming document $d$, represented as a vector, we compute a similarity metric $\Gamma_0 \colon \mathbb{R}^{k_1} \times \mathbb{R}^{k_1} \rightarrow \mathbb{R}$
	between the document vector and each of the existing centroids $\{ i \mid H(i, L(d)) \neq \bottom \}$.
	If the largest similarity exceeds a threshold $\tau$ for cluster index $j$, then we set $C(d) = j$. In that case, we also update the value of
	$H(i, L(d))$ to include new information from document $d$, as detailed below under ``$H$ update.'' If none of the similarity values exceed a threshold
	$\tau$, we find the first $i$ such that $H(i, L(d)) = \bottom$ (the first cluster id which is still unassigned), and set $C(d) = i$, therefore creating a new cluster. We again follow an ``$H$ update'' -- this time for starting a new cluster.

    In both cases, we also update the function $G$, by selecting the best crosslingual cluster for the recently updated (or created) monolingual cluster. To this end, we use another similarity metric $\Gamma_1 \colon \mathbb{R}^{k_2} \times \mathbb{R}^{k_2} \rightarrow \mathbb{R}$. Accordingly, we compute the similarity (using $\Gamma_1$) between the updated (or created) monolingual cluster and all monolingual clusters in each candidate crosslingual cluster, in the crosslingual feature space. The crosslingual cluster with highest sum of similarity scores is then selected. We also experimented computing this similarity by considering just the monolingual cluster of a particular ``pivot language''. The pivot language is a language that serves as the main indicator for a given crosslingual cluster. In our experiments, we mostly use English as the pivot language.

	\paragraph{$H$ Update}
	
	To update $H$, we maintain a centroid for each cluster that is created as the average of all monolingual representations of documents
	that belong to that cluster. This is done for each language separately. This update can be done in $O(k_1)$ time in each step. Similarly, the update of $G$ can be done in $O(k_2)$ time.
	In principle, we consider an ``infinite'' stream of documents, which means the number of documents in each cluster can be large.
	As such, for efficiency purposes, updates to $H$ are immutable, which means that when a document is assigned to a monolingual cluster, that assignment is never changed.

	\paragraph{$G$ Update}
	
	As described, updates to function $G$ result in associating a monolingual cluster with a crosslingual cluster (and consequently, other monolingual clusters). Therefore, errors committed in updating $G$ are of a higher magnitude than those committed in $H$, since they involve groups of documents. We also note that the best crosslingual cluster for a particular monolingual cluster might not be found right at the beginning of the process. We experiment with two types of updates to $G$. One which is immutable, in which changes to $G$ are not reversed (and are described above), and one in which we introduce a novel technique to make a sequence of changes to $G$ if necessary, as a mechanism to self-correct past clustering decisions. When a past decision is modified, it may result in a chaining of consequent modifications (``toppling of dominoes'') which need to be evaluated. We coin this method ``domino-toppling''.
	
	The motivation behind this technique is the change in news stories over time. The technique allows the method to modify past crosslingual clustering decisions and enables higher quality clustering results. When a past decision is modified, it may result in a chain of consequent modifications which need to be evaluated.

	\begin{figure}

\begin{footnotesize}
\framebox{\parbox{\columnwidth}{

{\bf Inputs:} A monolingual cluster $c$ and a list of pairs $\langle a_j, \Gamma_1(c, a_j) \rangle$, $j \in [N]$.

$\,$

{\bf Algorithm:}

\begin{itemize}[noitemsep,topsep=0pt,parsep=0pt,partopsep=0pt]

\item For all pairs $\langle a_j, \Gamma_1(c,a_j)\rangle$, $j \in [N]$, ordered by the second coordinate:
\begin{itemize}[noitemsep,topsep=0pt,parsep=0pt,partopsep=0pt]

\item If $L(c)$ is not in $a_j$, add $c$ to $a_j$ and break.
\item Otherwise, let $y \leftarrow M(a_j, L(c))$. If $\Gamma_1(c,a_j) > \Gamma_1(y, a_j)$ then:
\begin{itemize}[noitemsep,topsep=0pt,parsep=0pt,partopsep=0pt]

\item Add $c$ to $a_j$, remove $y$ from $a_j$ and call domino toppling with $y$ playing the role of $c$ and break.

\end{itemize}

\end{itemize}

\item If $c$ is left unassigned, create $a_{N+1}$ and add $c$ to it.

\end{itemize}

	}
	}
\end{footnotesize}

		\caption{Crosslingual ``domino-toppling''. $a_j$ is the $j$th crosslingual cluster (out of total $N$ clusters) and $\Gamma_1$ is the similarity between
them as in \S\ref{sec:doc}. $L(c)$ is the language for cluster $c$. $M(a, \ell)$ returns the monolingual cluster for language $\ell \in \mathcal{L}$ in crosslingual cluster $a$. See text for details. }
		\label{algorithm:domino_toppling}

	\end{figure}

\ignore{
	\begin{algorithm}[t]
		\SetKwProg{Fn}{Function}{:}{end}
		\Fn{domino-toppling ($c$)}{
			\nl $r(c) := $ list of pairs $\langle a_j, \Gamma_1(c, a_j)\rangle$\;
			\nl sort $r(c)$ by similarity $\Gamma_1(c, a_j)$\;
			\nl\ForEach{$\langle a_j, \Gamma_1(c, a_j)\rangle$ in $r(c)$}{
				
				\nl\uIf{$L(c)$ not in $a_j$}{
					\nl			add $c$ to $a_j$\;
					\nl			break\;
				}\uElse{
					\nl $y:= M(a_j, L(c))$\;
					\nl		\uIf{$\Gamma_1(c, a_j) > \Gamma_1(y, a_j)$}{
						\nl			add $c$ to $a_j$\;
						\nl			remove $y$ from $a_j$\;
						\nl			\textit{domino-toppling}($y$)\;
						\nl			break\;
					}{
					}
				}
			}
		}
		\caption{Crosslingual ``domino-toppling'' }
		\label{algorithm:domino_toppling}
	\end{algorithm}
}

	Our method of ``domino-toppling'' works by making (potentially sequences) of changes to previous clustering decisions for the crosslingual clusters, at each step placing a residual monolingual cluster in a crosslingual cluster that is most similar to it.
Figure~\ref{algorithm:domino_toppling} gives the pseudocode for domino toppling. 

	This ``domino-toppling'' technique could have in principle a quadratic complexity in the number of crosslingual clusters. However, we have verified that in practice it converges very fast, and in our evaluation dataset only 1\% of the crosslingual updates result in topples. We apply this technique only to update $G$ (and not $H$) because reversing cluster assignments in $G$ can be done much more efficiently than in $H$ -- the total number of monolingual clusters (the clustered elements in $G$) is significantly smaller than the number of documents (the clustered elements in $H$). \sebastiao{Crosslingual clustering is also a harder problem, which motivated the additional effort of developing this algorithm.}

	\section{Document Representation}
	\label{sec:doc}
	In this section, we give more details about the way we construct the document representations in the monolingual and crosslingual spaces. In particular, we introduce the definition of the similarity functions $\Gamma_0$ and $\Gamma_1$ that were referred in \S\ref{section:alg}.
	
	\paragraph{Monolingual Representation} The monolingual representation for each document $d$ in language $L(d)$ is a vector in $\mathbb{R}^{k_1}$ constructed from several TF-IDF subvectors with words,  word lemmas and named entities. Each subvector is repeated for different sections of the document, the title, the body and both of them together. Besides these text fields and document timestamps, no other metadata was used. To detect named entities, we used 
	Priberam's Text Analysis \cite{amaral2008priberam} for English and Spanish, and Turbo Parser \cite{Martins2013ACL} for German. The extracted entities consist of people, organizations, places and other types.
	
	
 	\paragraph{Crosslingual Representation} In the crosslingual space, a document representation is a vector in $\mathbb{R}^{k_2}$. Let $e(d,i)$ be a crosslingual embedding of word $i$ in the document $d$, which is a vector of length $m$. Then the document representation $v(d)$ of $d$ consists of subvectors of the form
	$v(d) = \sum_{i=1}^n t_i e(d,i)$, where $t_i$ is the TF-IDF score of the $i$th word in the relevant section of the document (title, body or both). As detailed further in \S\ref{section:experiments} we compute IDF values from a large pre-training dataset. Furthermore, for both the monolingual and crosslingual cases, we also experiment with using document timestamp features, as explained in \S\ref{section:similarity_metric}. We use a new set of diverse timestamp features in addition to the simple absolute difference (in hours) between timestamps used by \newcite{rupnik2016news}.

	\subsection{Similarity Metrics}
	\label{section:similarity_metric}
	Our similarity metric computes weighted cosine similarity on the different subvectors, both in the case of monolingual clustering and crosslingual clustering. Formally, for the monolingual case, the similarity is given by a function defined as: 
	
	\begin{equation}
	\Gamma_0(d_j, c_l) =  \sum_{i=1}^{K}\phi_i(d_j, c_l)\cdot q^0_i + \sum_{i=1}^3 \gamma_i(d_j, c_l) \cdot q^1_i. \label{eq:q}
	\end{equation}
	
	\noindent and is computed on the TF-IDF subvectors where $K$ is the number of subvectors for the relevant document representation. For the crosslingual case, we discuss below the function $\Gamma_1$, which has a similar structure.
	
	Here, $d_j$ is the $j$th document in the stream and $c_l$ is a monolingual cluster. The function $\phi_i(d_j, c_l)$ returns the cosine similarity between the document representation of the $j$th document and the centroid for cluster $c_l$. The vector $q^0$ denotes the weights through which each of the cosine similarity values for each subvectors are weighted, whereas $q^1$ denotes the weights for the timestamp features, as detailed further. Details on learning the weights $q^0$ and $q^1$ are discussed in~\S\ref{sec:fweights}.
	
	The function $\gamma(d,c)$ that maps a pair of document and cluster	to $\mathbb{R}^3$ is defined as follows. Let
	
	\begin{equation}\label{eq:timegauss}
	f(t) = \exp\left(-\displaystyle\frac{(t - \mu)^2}{2\sigma^2}\right)
	\end{equation}
	
	\noindent for a given $\mu$ and $\sigma > 0$.
	For each document $d$ and cluster $c$, we generate the following three-dimensional vector $\gamma(d,c) = (s_1, s_2, s_3)$:
	
	\begin{itemizesquish}{-0.3em}{0.5em}
		
		\item $s_1 = f(t(d) - n_1(c))$ where $t(d_j)$ is the timestamp for document $d$ and $n_1(c)$ is the timestamp for the newest document in cluster $c$.
		
		\item $s_2 = f(t(d) - n_2(c))$ where $n_2(c)$ is the average timestamp for all documents in cluster $c$.
		
		\item $s_3 = f(t(d) - n_3(c))$ where $n_3(c)$ is the timestamp for the oldest document in cluster $c$.
		
	\end{itemizesquish} 
	
	These three timestamp\footnote{Timestamps are given in hours since 1970.} features model the time aspect of the online stream of news data and help disambiguate clustering decisions, since time is a valuable indicator that a news story has changed, even if a cluster representation has a reasonable match in the textual features with the incoming document. The same way a news story becomes popular and fades over time~\cite{lerman2010using}, we model the probability of a document belonging to a cluster (in terms of timestamp difference) with a probability distribution.

	For the case of crosslingual clustering, we introduce $\Gamma_1$, which has a similar definition to $\Gamma_0$, only instead of passing document/cluster similarity feature vectors, we pass cluster/cluster similarities, across all language pairs. Furthermore, the features are the crosslingual embedding vectors of the sections title, body and both combined (similarly to the monolingual case) and the timestamp features. For denoting the cluster timestamp, we use the average timestamps of all articles in it.
	
	\subsection{Learning to Rank Candidates}
	\label{sec:fweights}
	In~\S\ref{section:similarity_metric} we introduced $q^0$ and $q^1$ as the weight vectors for the several document representation features. We experiment with both setting these weights to just $1$ ($q^0_i = 1 \, \forall i$ and $q^1_j=1 \, \forall j \in [3]$) and also learning these weights using support vector machines (SVMs). To generate the SVM training data, we simulate the execution of the algorithm on a training data partition (which we do not get evaluated on) and in which the gold standard labels are given. We run the algorithm using only the first subvector $\phi_1(d_j,c_l)$, which is the TF-IDF vector with the words of the document in the body and title. For each incoming document, we create a collection of positive examples, for the document and the clusters which share at least one document in the gold labeling. We then generate $20$ negative examples for the document from the $20$ best-matching clusters which are not correct. To find out the best-matching clusters, we rank them according to their similarity to the input document using only the first subvector $\phi_1(d_j,c_l)$.
	
	Using this scheme we generate a collection of ranking examples (one for each document in the dataset, with the ranking of the best cluster matches), which are then trained using the SVMRank algorithm \cite{joachims2002optimizing}. We run 5-fold cross-validation on this data to select the best model, and train both a separate model for each language according to $\Gamma_0$ and a crosslingual model according to $\Gamma_1$.

	\section{Experiments}
	\label{section:experiments}
	Our system was designed to cluster documents from a (potentially infinite) real-word data stream. 
	The datasets typically used in the literature (TDT, Reuters) have a small number of clusters ($\approx$ 20) with coarse topics (economy, society, etc.), and therefore are not relevant to the use case of media monitoring we treat - as it requires much more fine-grained story clusters about particular events.
	To evaluate our approach, we adapted a dataset constructed for the different purpose of binary classification of joining cluster pairs.\footnote{\url{https://github.com/rupnikj/jair_paper}} 
	We processed it to become a collection of articles annotated with monolingual and crosslingual cluster labels.\footnote{\sebastiao{The code and data we used} is available at \url{https://github.com/priberam/news-clustering}.}
	 
	Statistics about this dataset are given in Table~\ref{jair-table}. As described further, we tune the hyper-parameter $\tau$ on the development set. As for the hyper-parameters related to the timestamp features, we fixed $\mu = 0$ and tuned $\sigma$ on the development set, yielding  $\sigma = 72~\text{hours (3 days)}$.\footnote{This shows a relative robustness to reordering of the articles -- articles within 3 days of each other could appear anywhere in that window, and the algorithm would still perform well.} To compute IDF scores (which are global numbers computed across a corpus), we used a different and much larger dataset that we collected from Deutsche Welle's news website (\url{http://www.dw.com/}). The dataset consists of 77,268, 118,045 and 134,243 documents for Spanish, English and German, respectively. 

	The conclusions from our experiments are: (a) the weighting of the similarity metric features using SVM significantly outperforms
unsupervised baselines such as CluStream (Table~\ref{table:results}); (b) the SVM approach significantly helps to learn when to create a new cluster, compared to simple
grid search for the optimal $\tau$ (Table~\ref{table4}); (c) separating the feature space into one for monolingual clusters in the form of keywords and the other
for crosslingual clusters based on crosslingual embeddings significantly helps performance.

	\begin{table}
		\begin{center}
		{\small
		\centering
		\begin{tabular}{|l|l|cc|cc|}
			\hline  & Dataset & Size & Avg. L. & C & Avg. S. \\ \hline
			\multirow{3}{*}{\rotatebox{90}{train}} 
			& English & 12,233 & 434 & 593 & 21 \\
			&German & 4,043 &282  & 377 & 11  \\
			&Spanish & 4,527 & 355 & 416 & 11 \\
			\hline
			\multirow{3}{*}{\rotatebox{90}{test}} 
			& English & 8,726 & 521 & 222 & 39 \\
			&German & 2,101 & 440 & 118 & 18 \\
			&Spanish & 2,177 & 392 & 149 & 15 \\
			\hline
		\end{tabular}
		}
		\end{center}
		\caption{\label{jair-table} Statistics for the development and evaluation datasets, constructed from the dataset in \newcite{rupnik2016news}, as explained in \S\ref{section:experiments}. ``Size'' denotes the number of documents in the collection, ``Avg. L.'' is the average number of words in a document, ``C'' denotes the number of clusters in the collection and ``Avg. S.'' is the average number of documents in each cluster.}
	\end{table}
	
\paragraph{Evaluation Method}
	We evaluate clustering in the following manner: let $\mathrm{tp}$ be the number of correctly clustered-together document pairs,
	let $\mathrm{fp}$ be the number of incorrectly clustered-together document pairs and let $\mathrm{fn}$ be the number of incorrectly not-clustered-together
	document pairs. Then we report precision as $\frac{\mathrm{tp}}{\mathrm{tp}+\mathrm{fp}}$, recall as
	$\frac{\mathrm{tp}}{\mathrm{tp}+\mathrm{fn}}$ and F$_1$ as the harmonic mean of the precision and recall measures.

    We do the same to evaluate crosslingual clustering, but on a higher level: we count $\mathrm{tp}$, $\mathrm{fn}$ and $\mathrm{fp}$ for the decisions of clustering clusters, as crosslingual clusters are groups of monolingual gold clusters.

	\subsection{Monolingual Results}
	\label{section:monolingual}
	
	In our first set of experiments, we report results on monolingual clustering for each language separately.
	Monolingual clustering of a stream of documents is an important problem that has been inspected by others, such as by \newcite{ahmed2011unified} and by \newcite{aggarwal2006framework}. We compare our results to our own implementation of the online micro-clustering routine presented by \newcite{aggarwal2006framework}, which shall be referred to as CluStream. We note that CluStream of \newcite{aggarwal2006framework} has been a widely used state-of-the-art system in media monitoring companies as well as academia, and serves as a strong baseline to this day.

	In our preliminary experiments, we also evaluated an online latent semantic analysis method, in which
	the centroids we keep for the function $H$ (see \S\ref{section:alg})
	are the average of reduced dimensional vectors of the incoming documents as generated by an incremental singular
	value decomposition (SVD) of a document-term matrix that is updated after each incoming document.
	However, we discovered that online LSA performs significantly worse than representing the documents the way
	is described in \S\ref{sec:doc}. Furthermore, it was also significantly slower than our algorithm due to the time it took to perform singular value decomposition.\footnote{More specifically, we used an object of type \texttt{lsimodel} from the GenSim package that implements algorithms from \newcite{vrehuuvrek2010fast}. The GenSim package can be found at \url{https://pypi.python.org/pypi/gensim}.}

	\paragraph{Clustering experiments} Table~\ref{table:results} gives the final monolingual results on the three datasets. For English, we see that the significant improvement we get using our algorithm over the algorithm of \newcite{aggarwal2006framework} is due to an increased recall score. We also note that the trained models surpass the baseline for all languages, and that the timestamp feature (denoted by \textsc{TS}), while not required to beat the baseline, has a very relevant contribution in all cases. Although the results for both the baseline and our models seem to differ across languages, one can verify a consistent improvement from the latter to the former, suggesting that the score differences should be mostly tied to the different difficulty found across the datasets for each language. The presented scores show that our learning framework generalizes well to different languages and enables high quality clustering results.

		\begin{table}[t]
		\centering
		\begin{tabular}{|l|l|ccc|}
			\hline & algorithm &  F$_1$ &   P &  R \\ \hline
			\multirow{3}{*}{\rotatebox{90}{English}} 
			& CluStream & 79.0 & \textbf{98.6} & 65.9 \\
			& \small \textsc{Tokens}+\textsc{Lemmas}+\textsc{Ents} & 92.7 & 92.9 & \textbf{92.5} \\
			& $\,\,$ \small+\textsc{TS}  & \textbf{94.1} & 98.2 & 90.3 \\
			\hline
			\multirow{3}{*}{\rotatebox{90}{German}} 
			& CluStream & 89.7 & \textbf{99.9} & 81.3 \\
			& \small \textsc{Tokens}+\textsc{Lemmas}+\textsc{Ents} & 90.7 & 99.7 & 83.2 \\
			& $\,\,$ \small+\textsc{TS} &  \textbf{97.1}  & \textbf{99.9} & \textbf{94.5} \\
			\hline
			\multirow{3}{*}{\rotatebox{90}{Spanish}}
			& CluStream & 78.1 & 73.4 & 83.5 \\
			& \small \textsc{Tokens}+\textsc{Lemmas}+\textsc{Ents} & 88.8 & 95.9 & 82.7 \\
			& $\,\,$ \small+\textsc{TS} &  \textbf{94.2}  & \textbf{97.0} & \textbf{91.6} \\
			\hline
		\end{tabular}
		\caption{\label{table:results} Clustering results on the labeled dataset. We compare our algorithm (with and without timestamps) with the online micro-clustering routine of \newcite{aggarwal2006framework} (denoted by CluStream). The F$_1$ values are for the precision (P) and recall (R) in the following columns. See Table~\ref{table:ablation} for a legend of the different models. Best result for each language is in bold.}
	\end{table}
	
	To investigate the impact of the timestamp features, we ran an additional experiment using only the same three timestamp features as used in the best model on the English dataset. This experiment yielded scores of $F_1=61.1$, $P=44.5$ and $R=97.6$, which lead us to conclude that while these features are not competitive when used alone (hence temporal information by itself is not sufficient to predict the clusters), they contribute significantly to recall with the final feature ensemble. 
	
\begin{figure}[t]

	\includegraphics[width=3in]{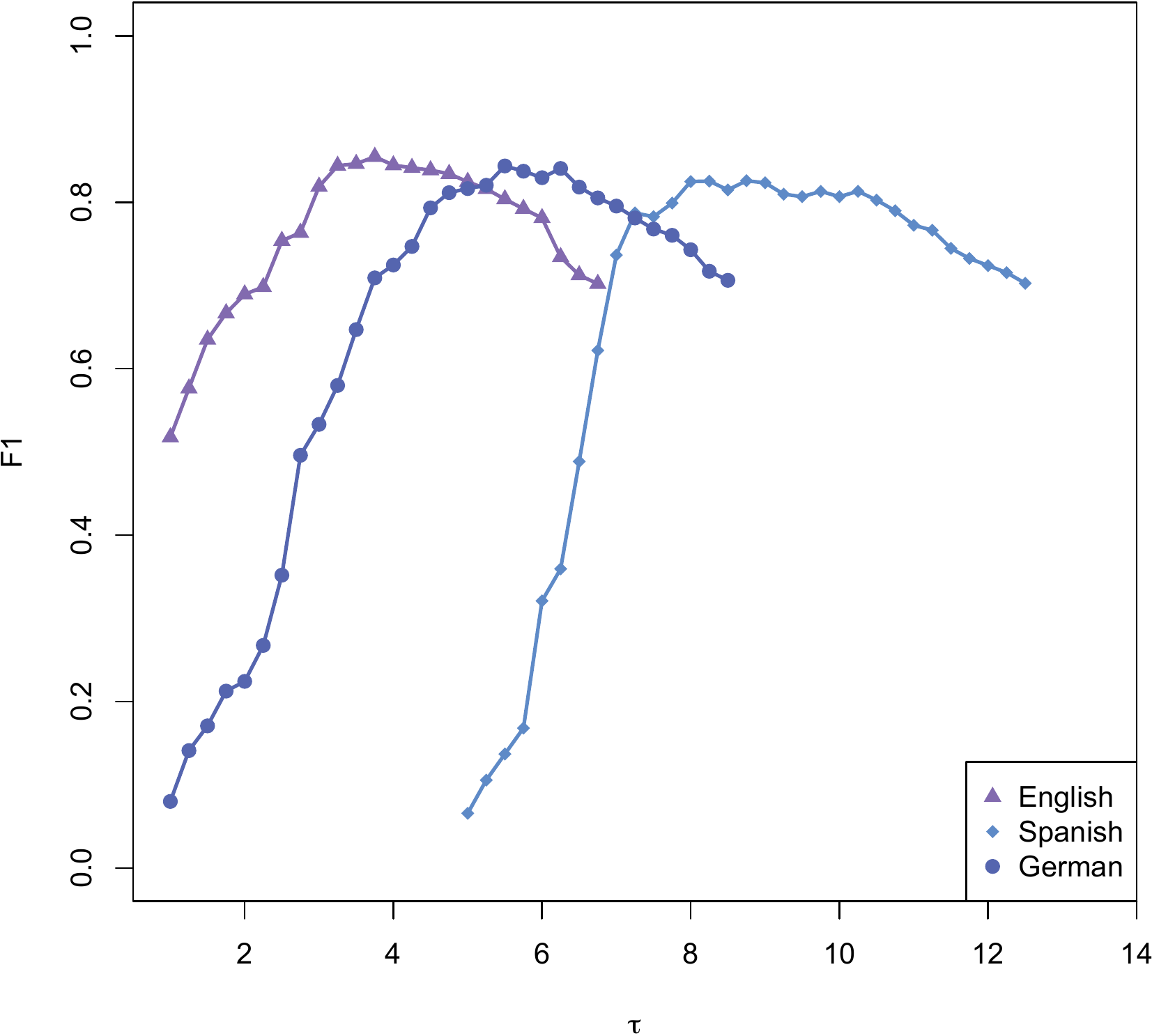}
	
	\caption{\label{fig:tau} The F$_1$ score of the different language development sets as a function of the threshold $\tau$. The first point for each language is identified using binary search.}
\end{figure}

	We note that as described in \S\ref{section:alg}, the optimization of the $\tau$ parameter is part of the development process.
The parameter $\tau$ is a similarity threshold used to decide when an incoming document should merge to the best cluster or create a new one. 
	We tune $\tau$ on the development set for each language, and the sensitivity to it is demonstrated in Figure~\ref{fig:tau} (this process is further referred to as $\tau_{\mathrm{search}}$). 
	Although applying grid-search on this parameter is the most immediate approach to this problem, we experimented with a different method which yielded superior results: as described further, we discuss how to do this process with an additional classifier (denoted SVM-merge), which captures more information about the incoming documents and the existing clusters.

	Additionally, we also experimented with computing the monolingual clusters with the same embeddings as used in the crosslingual clustering phase, which yielded poor results. In particular, this system achieved $F_1$ score of $74.8$ for English, which is below the bag-of-words baseline presented in Table~\ref{table:results}. This result supports the approach we then followed of having two separate feature spaces for the monolingual and crosslingual clustering systems, where the monolingual space is discrete and the crosslingual space is based on embeddings.

		\begin{table}
		\centering
		\begin{tabular}{|l|c|}
			\hline
			feature & accuracy \\
			\hline		
			\textsc{Tokens} & 85.5 \\
			\textsc{Tokens}+\textsc{Lemmas} & 85.9 \\
			\textsc{Tokens}+\textsc{Lemmas}+\textsc{Ents} & 86.5 \\
			\textsc{Tokens}+\textsc{Lemmas}+\textsc{Ents}+\textsc{TS} & 96.9 \\
			\hline
		\end{tabular}
		
		\caption{\label{table:ablation}Accuracy of the SVM ranker on the English training set. \textsc{Tokens} are the word token features, \textsc{Lemmas} are the lemma
			features for title and body, \textsc{Ents} are named entity features and \textsc{TS} are timestamp features. All features are described in detail in \S\ref{sec:doc}, and
			are listed for both the title and the body.}
		
	\end{table}

	\paragraph{SVM ranker experiments} To investigate the importance of each feature, we now consider in Table~\ref{table:ablation} the accuracy of the SVM ranker for English as described in \S\ref{section:similarity_metric}. We note that adding features increases the accuracy of the SVM ranker, especially the timestamp features. However, the timestamp feature actually interferes with our optimization of $\tau$ to identify when new clusters are needed, although they improve the SVM reranking accuracy. We speculate this is true because high accuracy in the reranking problem does not necessarily help with identifying when {\em new} clusters need to be opened. 
		\begin{table}[]
		\centering
		\begin{tabular}{|l|ccc|}
			\hline model & F$_1$ & P & R \\\hline
			$\tau_{\mathrm{search}}$ & 82.8 & 96.5 & 72.4 \\
			SVM-merge & 94.1 & 98.2 & 90.3 \\
			\hline
		\end{tabular}
		\caption{\label{table4}Comparison of two different cluster decision techniques for the English SVM model with all features (see Table~\ref{table:results}). The first method, $\tau_{\mathrm{search}}$, corresponds to executing grid-search to find the optimal clustering $\tau$ parameter (see \S\ref{section:alg}). SVM-merge is an alternative method in which we train an SVM binary classifier to decide if a new cluster should be created or not, where we use as
			features the maximal value of each coordinate for each document in a cluster. }
		\label{merge-model} 
	\end{table}
	To investigate this issue, we experimented with a different technique to learn when to create a new cluster. To this end, we trained another SVM classifier just to learn this decision, this time a binary classifier using LIBLINEAR~\cite{fan2008liblinear}, by passing the max of the similarity of each feature between the incoming document and the current clustering pool as the input feature vector. This way, the classifier learns when the current clusters, as a whole, are of a different news story than the incoming document. As presented in Table~\ref{merge-model}, this method, which we refer to as SVM-merge, solved the issue of searching for the optimal $\tau$ parameter for the SVM-rank model with timestamps, by greatly improving the F$_1$ score in respect to the original grid-search approach ($\tau_{\mathrm{search}}$). 
	
	\subsection{Crosslingual Results}
	
	As mentioned in \S\ref{section:alg}, crosslingual embeddings are used for crosslingual clustering. We experimented with the crosslingual embeddings of \newcite{gardner2015translation} and \newcite{ammar2016massively}.
	In our preliminary experiments we found that the former worked better for our use-case than the latter.
	
	\begin{table}[]
		\centering
		\begin{tabular}{|l|c|c|c|}
			\hline
			\multicolumn{1}{|c|}{crosslingual model} 	&  F$_1$ & P    & R    \\ \hline
            $\tau_{\mathrm{search}}$ (global)        & 72.7 & 89.8 & 61.0 \\ 
			$\tau_{\mathrm{search}}$ (pivot)      &84.0 & 83.0 &85.0\\ \hline
		\end{tabular}
	\caption{Crosslingual clustering results \sebastiao{when considering two different approaches to compute distances across crosslingual clusters} on the test set for Spanish, German and English. See text for details.}
	\label{crosswe-table} 
	\end{table}

	We test two different scenarios for optimizing the similarity threshold $\tau$ for the crosslingual case. Table~\ref{crosswe-table} shows the results for these experiments. First, we consider the simpler case of adjusting a global $\tau$ parameter for the crosslingual distances, as also described for the monolingual case. As shown, this method works poorly, since the $\tau$ grid-search could not find a reasonable $\tau$ which worked well for every possible language pair.
	
    Subsequently, we also consider the case of using English as a pivot language (see \S\ref{section:alg}), where distances for every other language are only compared to English, and crosslingual clustering decisions are made only based on this distance.\footnote{\sebastiao{In this case, all crosslingual clusters will have at least one pivot monolingual cluster, except for clusters which might stay unmerged as single-language degenerated crosslingual clusters.}}
    This yielded our best crosslingual score of $F_1{=}84.0$, confirming that crosslingual similarity is of higher quality between each language and English, for the embeddings we used. 
    This score represents only a small degradation in respect to the monolingual
    results, since clustering across different languages is a harder problem.

\section{Related Work}\label{section:related}

\sebastiao{Early research efforts, such as the TDT program~\cite{allan1998topic}, have studied news clustering for some time. The problem of online monolingual clustering algorithms (for English) has also received a fair amount of attention in the literature}. One of the earlier papers by \newcite{aggarwal2006framework} introduced a two-step clustering system with both offline and online components, where the online model is based on a streaming implementation of $k$-means and a bag-of-words document representation. Other authors have experimented with distributed representations, such as \newcite{ahmed2011unified}, who cluster news into {\em storylines} using Markov chain Monte Carlo methods, \newcite{rehurek_lrec} who used incremental Singular Value Decomposition (SVD) to find relevant topics from streaming data, and \newcite{sato2017distributed} who used the paragraph vector model~\cite{le2014distributed} in an offline clustering setting. 

More recently, crosslingual linking of clusters has been discussed by \newcite{rupnik2016news} in the context of linking existing clusters from the Event Registry \cite{leban2014event} in a batch fashion, and by \newcite{steinberger2016mediagist} who also present a batch clustering linking system. However, these are not ``truly'' online crosslingual clustering systems since they only decide on the linking of already-built monolingual clusters. In particular, \newcite{rupnik2016news} compute distances of document pairs across clusters using nearest neighbors, which might not scale well in an online setting. As detailed before, we adapted the cluster-linking dataset from \newcite{rupnik2016news} to evaluate our online crosslingual clustering approach. Preliminary work makes use of deep learning techniques \cite{xie2016unsupervised,guo2017improved} to cluster documents while learning
their representations, but not in an online or multilingual fashion, and with a very small number of cluster labels (4, in the case of the text benchmark).

In our work, we studied the problem of monolingual and crosslingual clustering, 
having experimented several directions and methods and the impact they have
on the final clustering quality. We described the first system which aggregates news articles into 
fine-grained story clusters across different languages in a completely online and scalable
fashion from a continuous stream.

\ignore{
	\section{Future Work}
	We propose two directions for future work:
	
	\begin{itemizesquish}{-0.3em}{0.5em}
		\item Training crosslingual embeddings jointly with the clustering ranking task. In particular, instead of learning word-level representations, document-level or cluster-level representations could yield better results.
		\item Instead of relying just on monolingual entity recognition, use crosslingual entity IDs (e.g., Wikipedia IDs) from an entity linking system and leverage them as additional crosslingual features, or as weak supervision.

	\end{itemizesquish}
}
	
	\section{Conclusion}
	
	We described a method for monolingual and crosslingual clustering of an incoming stream of documents. The method works by maintaining centroids for the monolingual and crosslingual clusters, where a monolingual cluster groups a set of documents and a crosslingual cluster groups a set of monolingual clusters. We presented an online crosslingual clustering method which auto-corrects past decisions in an efficient way. We showed that our method gives state-of-the-art results on a multilingual news article dataset for English, Spanish and German. Finally, we discussed how to leverage different SVM training procedures for ranking and classification to improve monolingual and crosslingual clustering decisions. Our system is integrated in a larger \textit{media monitoring} project \cite{E17-3029,germann2018summa} and solving the use-cases of monitors and journalists, having been validated with qualitative user testing.
	
	\section*{Acknowledgments}
	We would like to thank Esma Balk{\i}r, Nikos Papasarantopoulos, Afonso Mendes, Shashi Narayan and the anonymous reviewers for their feedback. This project was supported by the European H2020 project SUMMA, grant agreement 688139 (see \url{http://www.summa-project.eu}) and by a grant from Bloomberg.
	
\bibliography{nlp}

\begin{thebibliography}{21}
\expandafter\ifx\csname natexlab\endcsname\relax\def\natexlab#1{#1}\fi

\bibitem[{Aggarwal and Yu(2006)}]{aggarwal2006framework}
Charu~C. Aggarwal and Philip~S. Yu. 2006.
\newblock A framework for clustering massive text and categorical data streams.
\newblock In \emph{SDM}, pages 479--483. SIAM.

\bibitem[{Ahmed et~al.(2011)Ahmed, Ho, Eisenstein, Xing, Smola, and
  Teo}]{ahmed2011unified}
Amr Ahmed, Qirong Ho, Jacob Eisenstein, Eric Xing, Alexander~J Smola, and
  Choon~Hui Teo. 2011.
\newblock Unified analysis of streaming news.
\newblock In \emph{Proceedings of the 20th international conference on World
  wide web}, pages 267--276. ACM.

\bibitem[{Allan et~al.(1998)Allan, Carbonell, Doddington, Yamron, Yang
  et~al.}]{allan1998topic}
James Allan, Jaime Carbonell, George Doddington, Jonathan Yamron, Yiming Yang,
  et~al. 1998.
\newblock Topic detection and tracking pilot study: Final report.
\newblock In \emph{Proceedings of the DARPA broadcast news transcription and
  understanding workshop}.

\bibitem[{Amaral et~al.(2008)Amaral, Cassan, Figueira, Martins, Mendes, Mendes,
  Pina, and Pinto}]{amaral2008priberam}
Carlos Amaral, Ad{\'a}n Cassan, Helena Figueira, Andr{\'e} Martins, Afonso
  Mendes, Pedro Mendes, Jos{\'e} Pina, and Cl{\'a}udia Pinto. 2008.
\newblock Priberam's question answering system in {QA@ CLEF} 2008.
\newblock In \emph{Proceedings of the Workshop of the Cross-Language Evaluation
  Forum for European Languages}.

\bibitem[{Ammar et~al.(2016)Ammar, Mulcaire, Tsvetkov, Lample, Dyer, and
  Smith}]{ammar2016massively}
Waleed Ammar, George Mulcaire, Yulia Tsvetkov, Guillaume Lample, Chris Dyer,
  and Noah~A Smith. 2016.
\newblock Massively multilingual word embeddings.
\newblock \emph{arXiv preprint arXiv:1602.01925}.

\bibitem[{Fan et~al.(2008)Fan, Chang, Hsieh, Wang, and Lin}]{fan2008liblinear}
Rong-En Fan, Kai-Wei Chang, Cho-Jui Hsieh, Xiang-Rui Wang, and Chih-Jen Lin.
  2008.
\newblock Liblinear: A library for large linear classification.
\newblock \emph{Journal of machine learning research}, 9(Aug):1871--1874.

\bibitem[{Gardner et~al.(2015)Gardner, Huang, Papalexakis, Fu, Talukdar,
  Faloutsos, Sidiropoulos, and Mitchell}]{gardner2015translation}
Matt Gardner, Kejun Huang, Evangelos Papalexakis, Xiao Fu, Partha Talukdar,
  Christos Faloutsos, Nicholas Sidiropoulos, and Tom Mitchell. 2015.
\newblock Translation invariant word embeddings.
\newblock In \emph{Proceedings of EMNLP}.

\bibitem[{Germann et~al.(2018)Germann, Liepins, Barzdins, Gosko, Miranda, and
  Nogueira}]{germann2018summa}
Ulrich Germann, Renars Liepins, Guntis Barzdins, Didzis Gosko, Sebasti{\~a}o
  Miranda, and David Nogueira. 2018.
\newblock The summa platform: A scalable infrastructure for multi-lingual
  multi-media monitoring.
\newblock \emph{Proceedings of ACL 2018, System Demonstrations}.

\bibitem[{Guo et~al.(2017)Guo, Gao, Liu, and Yin}]{guo2017improved}
Xifeng Guo, Long Gao, Xinwang Liu, and Jianping Yin. 2017.
\newblock Improved deep embedded clustering with local structure preservation.
\newblock In \emph{Proceedings of {IJCAI}}.

\bibitem[{Joachims(2002)}]{joachims2002optimizing}
Thorsten Joachims. 2002.
\newblock Optimizing search engines using clickthrough data.
\newblock In \emph{Proceedings {ACM SIGKDD}}.

\bibitem[{Le and Mikolov(2014)}]{le2014distributed}
Quoc Le and Tomas Mikolov. 2014.
\newblock Distributed representations of sentences and documents.
\newblock In \emph{Proceedings of {ICML}}.

\bibitem[{Leban et~al.(2014)Leban, Fortuna, Brank, and
  Grobelnik}]{leban2014event}
Gregor Leban, Blaz Fortuna, Janez Brank, and Marko Grobelnik. 2014.
\newblock Event registry: learning about world events from news.
\newblock In \emph{Proceedings of {WWW}}.

\bibitem[{Lerman and Hogg(2010)}]{lerman2010using}
Kristina Lerman and Tad Hogg. 2010.
\newblock Using a model of social dynamics to predict popularity of news.
\newblock In \emph{Proceedings of {WWW}}.

\bibitem[{Liepins et~al.(2017)Liepins, Germann, Barzdins, Birch, Renals, Weber,
  van~der Kreeft, Bourlard, Prieto, Klejch, Bell, Lazaridis, Mendes, Riedel,
  Almeida, Balage, Cohen, Dwojak, Garner, Giefer, Junczys-Dowmunt, Imran,
  Nogueira, Ali, Miranda, Popescu-Belis, Miculicich~Werlen, Papasarantopoulos,
  Obamuyide, Jones, Dalvi, Vlachos, Wang, Tong, Sennrich, Pappas, Narayan,
  Damonte, Durrani, Khurana, Abdelali, Sajjad, Vogel, Sheppey, Hernon, and
  Mitchell}]{E17-3029}
Renars Liepins, Ulrich Germann, Guntis Barzdins, Alexandra Birch, Steve Renals,
  Susanne Weber, Peggy van~der Kreeft, Herve Bourlard, Jo{\~a}o Prieto, Ondrej
  Klejch, Peter Bell, Alexandros Lazaridis, Alfonso Mendes, Sebastian Riedel,
  Mariana S.~C. Almeida, Pedro Balage, Shay~B. Cohen, Tomasz Dwojak, Philip~N.
  Garner, Andreas Giefer, Marcin Junczys-Dowmunt, Hina Imran, David Nogueira,
  Ahmed Ali, Sebasti{\~a}o Miranda, Andrei Popescu-Belis, Lesly
  Miculicich~Werlen, Nikos Papasarantopoulos, Abiola Obamuyide, Clive Jones,
  Fahim Dalvi, Andreas Vlachos, Yang Wang, Sibo Tong, Rico Sennrich, Nikolaos
  Pappas, Shashi Narayan, Marco Damonte, Nadir Durrani, Sameer Khurana, Ahmed
  Abdelali, Hassan Sajjad, Stephan Vogel, David Sheppey, Chris Hernon, and Jeff
  Mitchell. 2017.
\newblock The {SUMMA} platform prototype.
\newblock In \emph{Proceedings of the Software Demonstrations of EACL}.

\bibitem[{Martins et~al.(2013)Martins, Almeida, and Smith}]{Martins2013ACL}
Andr\'{e} F.~T. Martins, Miguel~B. Almeida, and Noah~A. Smith. 2013.
\newblock Turning on the turbo: Fast third-order non-projective turbo parsers.
\newblock In \emph{Proceedings of {ACL}}.

\bibitem[{{\v{R}}eh{\r u}{\v{r}}ek(2010)}]{vrehuuvrek2010fast}
Radim {\v{R}}eh{\r u}{\v{r}}ek. 2010.
\newblock Fast and faster: A comparison of two streamed matrix decomposition
  algorithms.

\bibitem[{{\v R}eh{\r u}{\v r}ek and Sojka(2010)}]{rehurek_lrec}
Radim {\v R}eh{\r u}{\v r}ek and Petr Sojka. 2010.
\newblock {Software Framework for Topic Modelling with Large Corpora}.
\newblock In \emph{{Proceedings of the LREC 2010 Workshop on New Challenges for
  NLP Frameworks}}.

\bibitem[{Rupnik et~al.(2016)Rupnik, Muhic, Leban, Skraba, Fortuna, and
  Grobelnik}]{rupnik2016news}
Jan Rupnik, Andrej Muhic, Gregor Leban, Primoz Skraba, Blaz Fortuna, and Marko
  Grobelnik. 2016.
\newblock News across languages-cross-lingual document similarity and event
  tracking.
\newblock \emph{Journal of Artificial Intelligence Research}, 55:283--316.

\bibitem[{Sato et~al.(2017)Sato, Brockmeier, Kontonatsios, Mu, Goulermas,
  Tsujii, and Ananiadou}]{sato2017distributed}
Motoki Sato, Austin~J Brockmeier, Georgios Kontonatsios, Tingting Mu, John~Y
  Goulermas, Jun'ichi Tsujii, and Sophia Ananiadou. 2017.
\newblock Distributed document and phrase co-embeddings for descriptive
  clustering.
\newblock In \emph{Proceedings of {EACL}}.

\bibitem[{Steinberger(2016)}]{steinberger2016mediagist}
Josef Steinberger. 2016.
\newblock {MediaGist}: A cross-lingual analyser of aggregated news and
  commentaries.
\newblock In \emph{Proceedings of {ACL}}.

\bibitem[{Xie et~al.(2016)Xie, Girshick, and Farhadi}]{xie2016unsupervised}
Junyuan Xie, Ross Girshick, and Ali Farhadi. 2016.
\newblock Unsupervised deep embedding for clustering analysis.
\newblock In \emph{Proceedings of {ICML}}.

\end{thebibliography}
\bibliographystyle{acl_natbib}	
\end{document}